\documentclass[11pt]{article}
\usepackage{coling2020}
\usepackage{times}
\usepackage{url}
\usepackage{latexsym}
\usepackage{graphicx}
\usepackage{tabularx}
\usepackage[dvipsnames]{xcolor}
\usepackage{subcaption}
\usepackage{longtable}
\usepackage{array,multirow}
\usepackage{tabularx}
\usepackage{ltablex}
\usepackage{float}
\usepackage{color, soul}

\usepackage{hyperref}
\usepackage[dvipsnames]{xcolor}

\title{A Taxonomy of Empathetic Response Intents in Human Social Conversations}

\author{Anuradha Welivita and Pearl Pu\\
  School of Computer and Communication Sciences \\
  École polytechnique fédérale de Lausanne \\
  Switzerland \\
  {\tt \{kalpani.welivita,pearl.pu\}@epfl.ch} 
  \\}

\date{}

\begin{document}
\maketitle
\begin{abstract}
 
  Open-domain conversational agents or chatbots are becoming increasingly popular in the natural language processing community. One of the challenges is enabling them to converse in an empathetic manner. Current neural response generation methods rely solely on end-to-end learning from large scale conversation data to generate dialogues. This approach can produce socially unacceptable responses due to the lack of large-scale quality data used to train the neural models.  However, recent work has shown the promise of combining dialogue act/intent modelling and neural response generation. This hybrid method improves the response quality of chatbots and makes them more controllable and interpretable. A key element in dialog intent modelling is the development of a taxonomy. Inspired by this idea, we have manually labeled 500 response intents using a subset of a sizeable empathetic dialogue dataset (25K dialogues). Our goal is to produce a large-scale taxonomy for empathetic response intents. Furthermore, using lexical and machine learning methods, we automatically analysed both speaker and listener utterances of the entire dataset with identified response intents and 32 emotion categories. Finally, we use information visualization methods to summarize emotional dialogue exchange patterns and their temporal progression. These results reveal novel and important empathy patterns in human-human open-domain conversations and can serve as heuristics for hybrid approaches.

\end{abstract}

\section{Introduction}
\label{sec:introduction}

\blfootnote{
    %
    %
    %
    %
     \hspace{-0.65cm}  
     This work is licensed under a Creative Commons 
     Attribution 4.0 International Licence.
     Licence details:
     \url{http://creativecommons.org/licenses/by/4.0/}.
    %
    %
}

Inspired by the recent success of deep neural networks for natural language processing (NLP) tasks such as language modeling \cite{Recurrent_neural_network_based_language_model} and machine translation \cite{Sequence_to_sequence_learning_with_neural_networks}, neural response generation is currently at the forefront of research in the NLP community. Recent advances in this field have proven the efficacy of deep neural networks in modelling both task-oriented and open-domain dialogue systems \cite{semantically_conditioned_lstmbased_natural_language_generation_for_spoken_dialogue_systems,Sequence_to_sequence_learning_with_neural_networks,encoder-decoder}. Most of the existing neutral conversation models are capable of generating syntactically and contextually well-formed responses. Some of the work also focuses on enabling chatbots to generate emotionally colored and affect-rich responses \cite{asghar,ecm,meed}. Despite the efforts in modeling affect in natural language, work that focuses specifically on modeling empathy in chatbots is relatively limited and remains an open research question \cite{empathic_response_generation_in_chatbots}.

Empathy plays a vital role in human psychological processes for smooth social interaction \cite{The_Neurodevelopment_of_Empathy_in_Humans}. Empathy-related responding includes caring and sympathetic concerns for other people. Humans are born with core {\em affect} neural circuitry, and they gradually develop the ability to apprehend the emotional states of others and respond in an empathetic manner. Empathy motivates pro-social behavior and increases the sense of social bonding \cite{Empathic_responding_sympathy_and_personal_distress}. Therefore, in the context of social interaction, a chatbot needs to be empathetic to maintain healthy interaction with humans and develop trust. The task of augmenting social chatbots with empathy is challenging because the generated responses have to be appropriate in terms of both content and emotion information \cite{empathic_response_generation_in_chatbots}. 

Several neural response generation models have attempted to address this challenge in a fully data-driven manner. For example, Rashkin et al. \shortcite{ed}, use the full transformer architecture \cite{encoder-decoder} pre-trained on $1.7$ billion Reddit conversations and fine-tuned on the EmpatheticDialogues dataset \cite{ed} to generate empathetic responses. Lin et al. \shortcite{caire} adapt the Generative Pretrained Transformer (GPT) \cite{Improving_language_understanding_by_generative_pre-training} to empathetic response generation task by fine-tuning it on the PersonaChat \cite{Personalizing_dialogue_agents} and EmpatheticDialogues datasets. Even though these models are capable of mimicking human empathetic conversation patterns in some ways,  it is often unpredictable what the chatbots might generate, for example, they may generate inconsiderate remarks, redundant responses, asking the same questions repeatedly, or any combinations of them. Since it is really important to respond to humans' emotions appropriately, we believe controllability of response generation is essential.

Several other neural response generation approaches attempt to gain control over the generated response by conditioning it on a manually specified emotion label \cite{ecm,mojitalk,tone_aware,Generating_Responses_with_a_Specific_Emotion_in_Dialog} or using affective loss functions based on heuristics such as minimizing or maximizing affective dissonance between prompts and responses \cite{asghar}. These models claim to generate emotionally more appropriate responses than those generated from purely data-driven models. However, the primary concern of these handcrafted rules is its practicality. No prior work has shown normative associations between the speaker's emotions and the corresponding listener's emotions. As our work would reveal, listeners are much more likely to respond to sad or angry emotions with questioning than expressing similar or opposite emotions in the first turn. Xu et al. \shortcite{microsoft}, however, has shown the benefit of incorporating dialogue acts as policies in designing a social chatbot. They were able to avoid the need to manually condition the next response with a label by jointly modeling dialogue act selection and response generation. Their framework first selects a dialogue act from a policy network according to the dialogue history. The generation network then generates a response based on both dialogue history and the selected dialogue act. It is thus possible to explicitly learn human-human conversational patterns in social chitchat and generate more controlled and interpretable responses. Unfortunately, they did not study empathetic response generation.

\begin{figure}[ht!]
    \centering
        \includegraphics[width=0.7\textwidth]{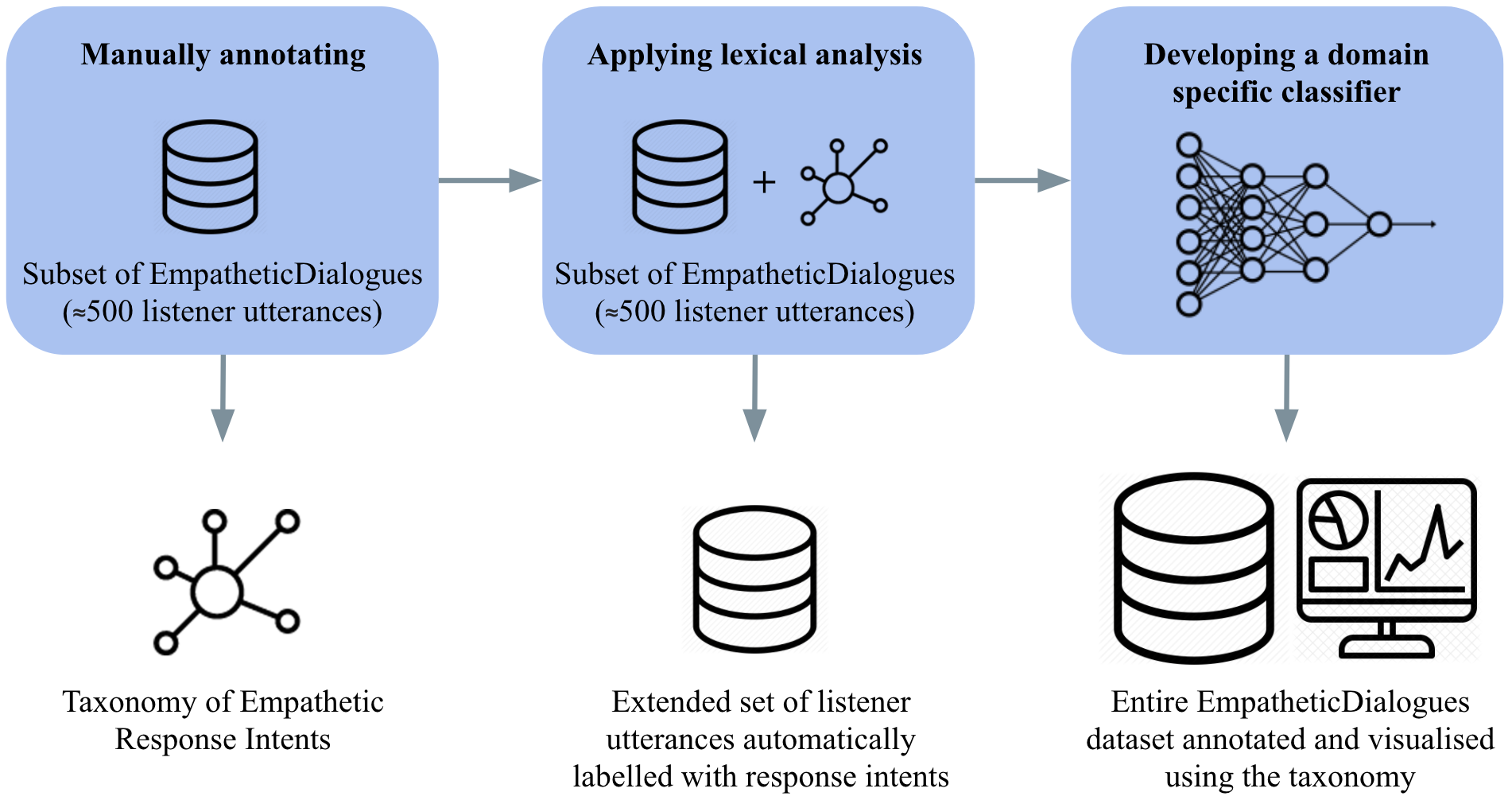}
    \caption{Three development steps for constructing the taxonomy of empathetic response intents.}
    \label{fig:scheme}
\end{figure}

To fill this gap, we have developed a taxonomy of empathetic listener intents by manually annotating around 500 utterances of the EmpatheticDialogues dataset \cite{ed}, covering 32 types of emotion categories. In the following, we first describe in detail how this taxonomy was derived (Figure \ref{fig:scheme}) and how we chose the dataset to support this annotation work. To extend this subset, we employ automatic techniques to label all speaker and listener utterances, covering 25k empathetic human-human conversations. To be able to explain the patterns and trends of the conversation flow, we employ visualization methods to illustrate the most frequent exchanges and reveal how they temporally vary as dialogues proceed. Finally, we discuss how these results can be used to derive more informed heuristics for controlling the neural response generation process.\footnote{Our source code and results are available at \url{https://github.com/anuradha1992/EmpatheticIntents}.}

\section{Related Work}
\label{sec:related_work}

To provide a background for this research, we begin by describing some of the existing theories related to empathy in other fields such as psychology and neuroscience and their limitations in incorporating them into the design of social chatbots. Then we describe seminal work on existing neural-based open-domain response generation systems and means by which they control the generated response. These studies serve as the motivation and inspiration for our work. Next, we discuss some existing dialogue-act/intent taxonomies and their limitations in modeling empathy in human social conversations.

\subsection{Theories of Empathy in Psychology and Neuroscience}\label{psychology}

Zillmann \shortcite{zillmann} defines empathy as a social emotion in response to the emotions of others. Further, he states that the evoked empathetic reaction itself constitutes an emotional experience, primarily because it is associated with increased excitement and awareness. It tends to be a \textit{feeling with} or \textit{feeling for} the observed party. Two of the most famous theories in the psychological literature to explain the phenomenon of empathy are the ``simulation theory" \cite{simulation} and the ``theory-theory" \cite{theory}. The ``simulation theory" states that a person understands another or empathizes by imagining himself in the other’s situation and seeing it from his perspective. The ``theory-theory" states that the ability to understand what another person is feeling is based on rules for how one should think and feel. 

A recent work by Singer and Klimecki \shortcite{empathy_and_compassion} in the field of Neuroscience states that empathy refers to the general capacity of humans to resonate with others' emotional states irrespective of their valence. However, when empathizing, they suggest that one should not confuse oneself with the other; i.e., one should still know that the emotion he resonates with is the emotion of another. The failure to separate that can lead to empathetic distress. According to them, the desirable way of empathizing with others is having compassion, which is a feeling of concern for another person's suffering accompanied by the motivation to help. However, all this work does not describe in detail specific means through which humans show empathy, especially via natural language dialogues. Also, most of these studies on empathic states focus on reactions to negative rather than positive events \cite{Modeling_Empathy_and_Distress_in_Reaction_to_News_Stories}. Hence, empathy for positive events remains less understood. In our study, we explore the means through which humans empathize with others both in positive and negative scenarios. 

\subsection{Neural Response Generation}\label{models} 

Xie et al. \shortcite{meed}, describe an end-to-end Multi-turn Emotionally Engaging Dialog model (MEED), capable of recognizing emotions and generating emotionally appropriate and human-like responses. Their GRU based Seq2Seq dialogue model consists of a hierarchical mechanism to track the conversation history in multi-turn dialogues combined with an additional emotion RNN to process the emotional information in each history utterance. They model affect exchanges in human dialogues using a dedicated embedding layer. This emotion recognition step enables the model to produce more emotionally appropriate responses for a given context. But their approach is entirely data-driven and lacks control and interpretability over generated responses. 

Chen et al. \shortcite{neutral_response_generation_with_relevant_emotions_for_short_text_conversation}, propose a model that can generate comments to posts in social media so that they are not only relevant in topic but also in emotion. To fully understand how emotions are expressed in conversations, they first analyse NTCIR-12 STC-1 collection \cite{ntcir}, a social-media conversation dataset. The results show that for posts with different emotions, the distributions of comment emotions are very different from each other, and only several emotions are appropriate for responding to a given post. Inspired by the findings, they extend the basic encoder-decoder neural network architecture \cite{encoder-decoder} with an RNN-based response emotion estimator, which takes in a post and estimates how relevant an emotion is for responding to the post. This information is fed into the decoder when generating the response. In this, the classifier automatically determines the emotion of the response. Xu et al. \shortcite{microsoft} incorporate dialogue acts as policies in their open-domain neural response generation model by performing learning with human-human conversations tagged with a dialogue act classifier. They jointly model dialogue act selection and response generation using a GRU based neural network consisting of a policy network and a generation network. The policy network first selects a dialogue act according to the conversation history, and then the generation network generates a response based on the conversation history and the selected dialogue act. They claim that with dialogue acts, they not only achieve significant improvement over response quality for a given context but also can explain why such achievements are possible. The above work motivated us to develop explicit empathetic response intents from the dataset. We believe they can inform the development of empathetic social chatbots by providing more control to and interpretation of the responses generated and render human-machine conversations more natural and engaging. 

\subsection{Dialogue-Act/Intent Taxonomies}\label{texonomies}

Work has been conducted to establish dialogue act/intent taxonomies both by analysing human-human and human-machine conversation datasets. Stolcke et al. \shortcite{Dialogue_Act_Modeling_for_Automatic_Tagging_and_Recognition_of_Conversational_Speech} propose a taxonomy of 42 mutually exclusive dialogue acts with the intention of enabling computational dialogue act modeling for conversational speech. They follow the standard Dialog Act Markup in Several Layers (DAMSL) tag set \cite{damsl} and modify it in several ways so they can easily distinguish utterances in conversational speech. Using this taxonomy, they produce a large hand-labeled database of 1,155 conversations from the Switchboard corpus of spontaneous human-to-human telephone conversations, which is widely used to train and test dialogue act classifiers. Montenegro et al. \shortcite{montenegro} propose a dialogue act taxonomy for a task-oriented virtual coach designed to improve the lives of the elderly. It is a multi-dimensional hierarchical taxonomy comprising of topic, intent, polarity, and entity labels at the top, in which the intent label classifies the utterance in classes related to the user’s communicative intentions such as `question', `inform', and `agree'. They use the taxonomy to manually annotate the user turns in 384 human-machine dialogues collected from a group of elderly. It aims to help the dialogue agent to detect goals, realities, obstacles, and ways forward of the particular topics the agent is designed to deal with. 

Existing dialogue-act/intent taxonomies are either too general as they were constructed for open-domain conversations or too specific as they were constructed for specific task-oriented scenarios. These taxonomies do not necessarily model empathy in human social conversations. Also, the above approaches do not use automatic approaches to extend the manual annotations, which make their datasets comparatively smaller. In our study, we present how a smaller set of human labeled sentences can be extended using lexical methods and use it to train a classifier to automatically annotate a larger corpus.


\section{Dataset}
\label{sec:dataset}

Many open-domain conversation datasets are publicly available mainly to assist tasks such as neural dialogue generation. Out of them, some datasets are multi-modal (e.g. IEMOCAP \cite{iemocap}, SEMAINE \cite{semaine}, MELD \cite{meld}) containing visual, acoustic and textual signals. Since they contain a lot of back-channel communication through facial expressions and speech tones, the text may not fully represent the contextual expression of intent. Datasets containing dialogues extracted from social media platforms such as Twitter (e.g., the Twitter Dialog Corpus \cite{twitter}) are often noisy, short, and different from real-world conversations and may contain a lot of toxic responses rather than compassionate ones. Also, datasets containing TV or movie transcripts (e.g., Emotionlines \cite{emotionlines}, OpenSubtitles \cite{os_2018}) and telephone recordings (e.g. Switchboard corpus \cite{Dialogue_Act_Modeling_for_Automatic_Tagging_and_Recognition_of_Conversational_Speech}) are a translation of voice into text, which does not fully model interactions that happen only through text. Even purely text-based daily conversation datasets such as DailyDialog \cite{dailydialog} are not guaranteed to contain empathetic responses. 

Rashkin et al. \shortcite{ed} introduced the EmpatheticDialogues dataset consisting of 24,856 open-domain, human-human conversations as a benchmark dataset to train and evaluate dialogue systems that can converse in an empathetic manner. Each conversation in this dataset is based on a situation associated with one of 32 emotions, which are selected from multiple annotation schemes, ranging from basic emotions derived from biological responses \cite{ekman,plutchik} to larger sets of subtle emotions derived from contextual situations \cite{skerry}. The dialogues are collected using ParlAI \cite{miller}, integrated with Amazon Mechanical Turk (MTurk), recruiting 810 US workers. During construction, the workers were instructed to show empathy when responding to conversations initiated by their speaker counterparts. Since almost all the dialogues in this dataset are empathetic, purely text-based, and most of which do not contain any toxic responses, we chose it to derive our taxonomy. An example conversation from this dataset is given in Table \ref{table:example}. Table \ref{table:statistics} shows the basic statistics of the dataset. The average number of turns per dialogue is close to $4$. The maximum number of dialogue turns in the dataset is $8$. However, not many dialogues exceed $4$ turns. Close to $77\%$ of the total number of dialogues contain only up to $4$ turns, and only $1.4\%$ of the dialogues contain up to $8$ turns.

\begin{table}
\begin{tabularx}{\textwidth}{|r X|}
\hline
Label: &  Afraid\vspace{0.5mm}\\
Situation: & Speaker felt this when... \textit{``I’ve been hearing noises around the house at night"} \vspace{0.5mm}\\ 
Conversation: & Speaker: \textit{I’ve been hearing some strange noises around
the house at night.}

Listener: \textit{oh no! That’s scary! What do you think it is?}

Speaker: \textit{I don’t know, that’s what’s making me anxious.}

Listener: \textit{I’m sorry to hear that. I wish I could help you
figure it out}\\

\hline
\caption{Example conversation taken from the EmpatheticDialogues dataset.}
\label{table:example}
\end{tabularx}
\end{table}



\begin{table}[ht!]
\centering
\begin{tabular}{|l|r|}
\hline Criteria
& Statistics \\ \hline

Total no. of dialogues & $24,856$ \\
Total no. of dialogue turns & $107,247$ \\
Average no. of turns per dialogue & $4.31$\\
Maximum no. of turns per dialogue  & $8$ ($345$ dialogues)\\
Minimum no. of turns per dialogue  & $1$ ($3$ dialogues)\\
Total no. of speaker turns  & $55,984$\\
Total no. of listener turns & $51,263$\\
Average no. of speaker tokens per dialogue turn & $17.88$\\
Average no. of listener tokens per dialogue turn & $13.69$\\

\hline
\end{tabular}
\caption{Statistics of the EmpatheticDialogues dataset used for analysis.}\label{table:statistics}
\end{table}

\vspace{-0.5cm}

\section{Taxonomy of Empathetic Response Intents}
\label{sec:taxonomy}

To develop the taxonomy, we investigated which intents are frequently associated with listeners when responding to different emotional situations in EmpatheticDialogues. We took a subset of the dataset with situations associated with the Plutchik’s 8 basic emotions \cite{plutchik} (joyful, anticipating, trusting, surprised, angry, afraid, sad, and disgusted), and manually analysed it to derive the listener intents associated with each type of emotions. In this process, $20$ dialogues belonging to each emotion were randomly selected and each sentence in all listener utterances were manually annotated by an expert evaluator with a label that best describes their intent. This resulted in $521$ sentences manually annotated with intent labels. Because an utterance can have multiple sentences, we decided to annotate each sentence in a listener's utterance with a unique intent label. For example, the two sentences comprising the utterance \textit{``Those symptoms are scary! Do you think it's Corona?"} would be annotated with separate intent labels ``Acknowledging" and ``Questioning", respectively. After analysing their occurrences and whether some of the intents can be grouped into a common intent, we were able to come up with a taxonomy of $15$ empathetic response intents. Table \ref{table:taxonomy_intents} presents this taxonomy with corresponding examples and occurrence frequencies. Words and phrases that were most helpful in annotating these examples with their corresponding intents are underlined. Manual annotation of empathetic response intents was carried out with reference to the context preceding an utterance. This way, we were able to distinguish utterances using similar words in the same order depending on their context. For example, sentences such as \textit{``I hope they find a vaccine soon."} can be categorised into two different intents, ``Encouraging" and ``Consoling" depending on whether the sentence follows a positive or negative emotional context, respectively.



\begin{tabularx}{\textwidth}{|X|X|r|}


\hline Category
& Examples & Frequency
\\ \hline

1. Questioning (to know further details or clarify)  & - \textit{\underline{What} are you looking forward to\underline{?}} & $24.38\%$\\

2. Acknowledging (Admitting as being fact) & - \textit{That \underline{sounds like} double good news. It was \underline{probably fun} having your hard work rewarded.} & $22.46\%$\\

3. Agreeing (Thinking/Saying the same) & - \textit{That's a great feeling, \underline{I agree!}} & $9.60\%$\\

4. Consoling  & - \textit{\underline{I hope} he gets the help he needs.} & $7.87\%$
\\

5. Encouraging & - \textit{\underline{Hopefully} you will catch those great deals!} & $5.37\%$\\

6. Sympathizing (Express feeling pity or sorrow for the person in trouble) & - \textit{So \underline{sorry to hear} that.} & $5.37\%$\\

7. Wishing  & - \textit{Hey... \underline{congratulations} to you!} & $4.41\%$ \\

8. Suggesting  & - \textit{\underline{Maybe} you two \underline{should} go to the pet store to try and find a new dog for him!} & $4.03\%$
\\

9. Sharing own thoughts/opinion & \textit{\underline{I would} love to have a boy too, but I'm not sure if I want another one or not.} & $4.03\%$\\

10. Sharing or relating to own experience & \textit{I had a friend who \underline{went through the same thing.}} & $3.84\%$\\

11. Advising & \textit{\underline{Don’t} take too much money with you.} & $2.69\%$\\

12. Expressing care or concern & \textit{\underline{I hope} the surgery went successfully and with no hassle.} & $2.30\%$\\

13. Expressing relief & \textit{\underline{Phew.. That’s a relief.}, I am glad you were okay.} & $1.53\%$\\

14. Disapproving & \textit{\underline{But} America is so great now! look at all the great things that are happening.} & $1.15\%$\\

15. Appreciating & \textit{\underline{You are very trusting.} It's nice to have a friend like you.} & $0.95\%$\\\hline

\caption{Taxonomy of empathetic response intents with corresponding examples and occurrence frequencies based on the manually annotated 521 listener utterances in the EmpatheticDialogues dataset.}
\label{table:taxonomy_intents}
\end{tabularx}

\section{Automatic Labelling of EmpatheticDialogues Using the Taxonomy}
\label{sec:labelling}

\subsection{Annotation Procedure}
\label{sec:annotation}

To annotate all the speaker and listener utterances in the EmpatheticDialogues dataset with emotion labels and response intents, we trained a BERT transformer-based classifier, as suggested by Devlin et al. \shortcite{bert}. Prior to selecting BERT as the classifier, we trained and tested a FastText classifier on the annotation task, but its accuracy was lower compared to BERT. We proceeded with the $8$ most frequent intents (questioning, acknowledging, agreeing, consoling, encouraging, sympathizing, wishing, and suggesting) in our taxonomy of empathetic listener intents and the $32$ types of emotion categories given in the EmpatheticDialogues dataset. The rest of the listener intents were classified as `neutral' since the emotion behind those intents were more on the neutral side. To expand the training data collected by manual annotation, we searched through the rest of the dataset using n-grams that are most indicative of the intent categories. For example, n-grams such as `100 \%', `absolutely', `definitely', `i agree', `me neither', `me too', and `i completely understand' are indicative of the intent `agreeing' and were used to collect more example utterances corresponding to that category. The most indicative n-grams used to collect more utterances for each of the intents are listed in Appendix A. 

During training, we initialized the representation network with weights from the pre-trained language model, RoBERTA \cite{roberta}, and fine-tuned the model on situation descriptions given in the EmpatheticDialogues dataset tagged with $32$ emotions and listener utterances tagged with 8 out of 15 intents from our taxonomy of empathetic response intents. The training, validation, and test sets comprised of $25 023$, $3 544$ and $3 225$ sentences respectively, which spanned equally across all emotion and intent categories. We trained the model with a peak learning rate of $2e^{-5}$ and a batch size of $32$ for $10$ epochs and obtained the classifier giving the lowest validation loss. The top-1 accuracy of our classifier with 41 labels over the test set was $65.88\%$, which is significantly higher than the accuracy of FastText \cite{fasttext} and DeepMoji \cite{deepmoji} classifiers trained on 32 emotion labels in the EmpatheticDialogues dataset. The latter two were considered as the state-of-the-art at that time, and achieved $43\%$ and $48\%$ accuracy on the EmpatheticDialogues test set, respectively \cite{ed}.

\subsection{Analysis of Emotion-Intent Exchange Patterns}

\begin{figure}[ht!]
    \centering
        \includegraphics[width=0.6\textwidth]{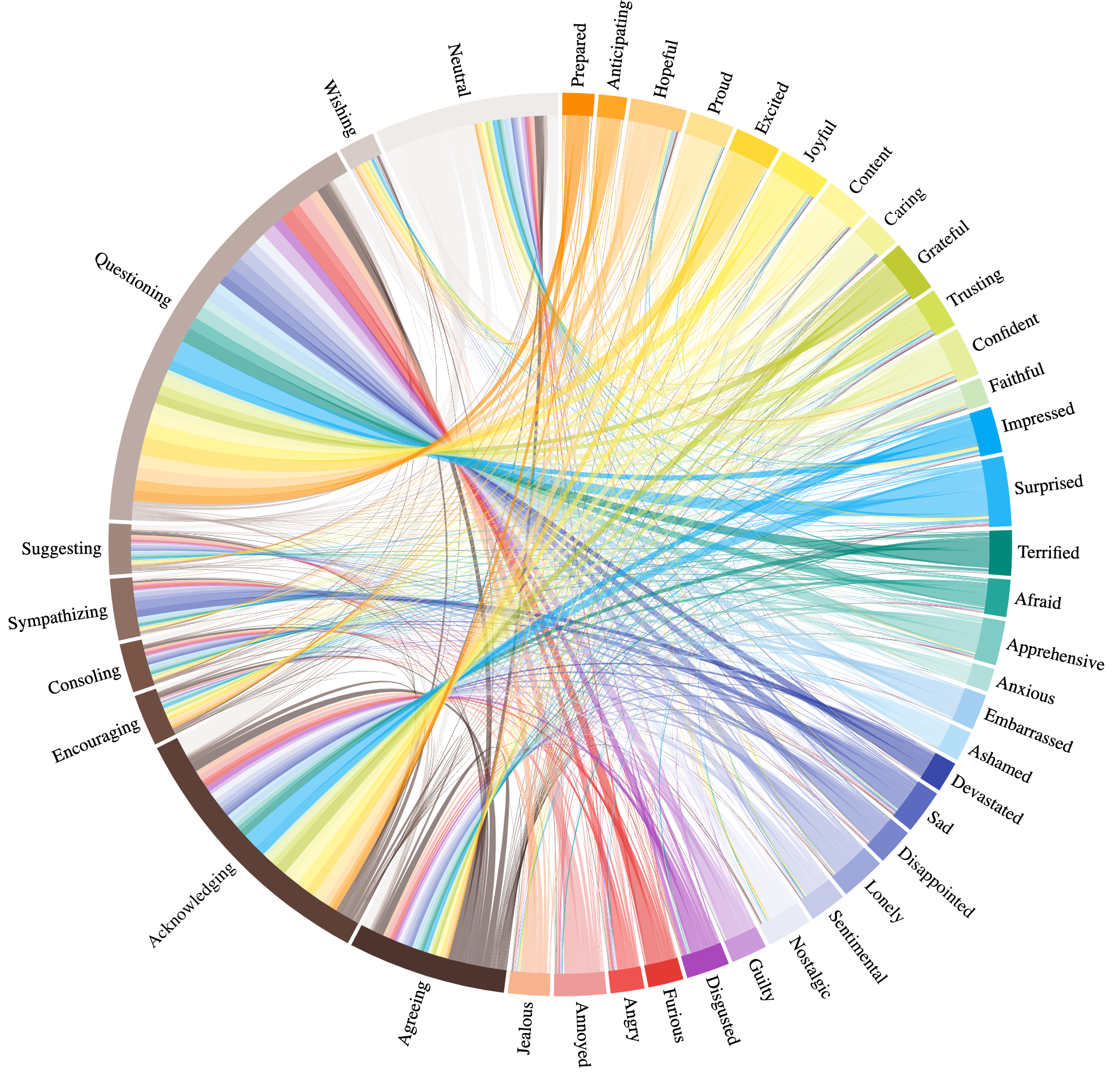}
    \caption{Visualization of emotion-intent exchanges between speakers and listeners in the EmpatheticDialogues dataset irrespective of the dialogue turn. Each chord connects co-occurring emotion-intent pairs. The chord leaving a particular arc represents the speaker's emotion or intent and gets connected to the arc representing the listener's emotion or intent that immediately follows in a conversation.}
    \label{fig:chord_diagram}
\end{figure}

Based on the above annotations, we analysed the most frequent response intents corresponding to different emotions expressed by speakers. In Figure \ref{fig:chord_diagram}, we visualize the emotion-intent exchanges taking place between speakers and listeners in the EmpatheticDialogues dataset. In this, each chord connects emotion-intent pairs that co-occur together. The chord leaving a particular arc represents the speaker's emotion or intent and gets connected to the arc representing the listener's emotion or intent that immediately follows. It can be seen that a significant proportion of speaker utterances contain a particular emotion in the 32 different emotion categories defined in EmpatheticDialogues, while most of the listener utterances contain a particular intent defined in our taxonomy. Instead of conveying a particular emotion, the listeners show their empathy via specific means described in our taxonomy. And the proportions of the arcs for each intent resembles the frequencies in the manually annotated subset. It serves as a validation that our taxonomy is indeed true for listener responses as it was applied to the entire dataset. 

It can be seen that `questioning' and `acknowledging' play a significant role in empathetic responses irrespective of the speaker's emotion---whether it is subtle or intense or has a positive or negative valence. Questioning enables the listener to sound more attentive and show interest in what the speaker describes. It prevents listeners from arriving at early conclusions, without knowing the situation in detail. It is also important to let speakers know that they have the right to feel the way they feel, even though listeners may not completely agree with their choices. Expressions of `acknowledgment' serve this purpose. This type of emotional interaction allows the speaker to elaborate on his feelings and what he is going through and feel validated at the same time. It can also be seen that some listener intents such as `encouraging' and `wishing' are frequently associated with positive speaker emotions, and some intents such as `sympathizing' and `consoling' are frequently associated with negative speaker emotions. A list of example utterance-response pairs corresponding to some of the most frequent emotion-intent exchanges ($\geq100$ times out of $\approx50$k utterance-response pairs in EmpatheticDialogues) are given in Appendix B. 

Next, we analysed how emotions and response intents shift over different turns in the dialogue as the dialogues progress in time. In this analysis, we discovered the most frequent emotion-intent flows that occur between speakers and listeners from the start to the end of conversations. To visualize the shift in emotions and intents over different dialogue turns, we computed the frequency of emotion-intent flow patterns up to $4$ dialogue turns and plotted the ones having a frequency $\geq 5$. The reason for selecting only the first $4$ turns in the dialogues is the fact that close to $77\%$ of the dialogues in the EmpatheticDialogues dataset contain up to $4$ turns and from this only $1.8\%$ of the dialogues go up to $8$ turns. Since there is comparatively much fewer data over dialogue turns from 5 to 8, we omitted these turns in our analysis.

\begin{figure}[ht!]
    \centering
        \includegraphics[width=0.95\textwidth]{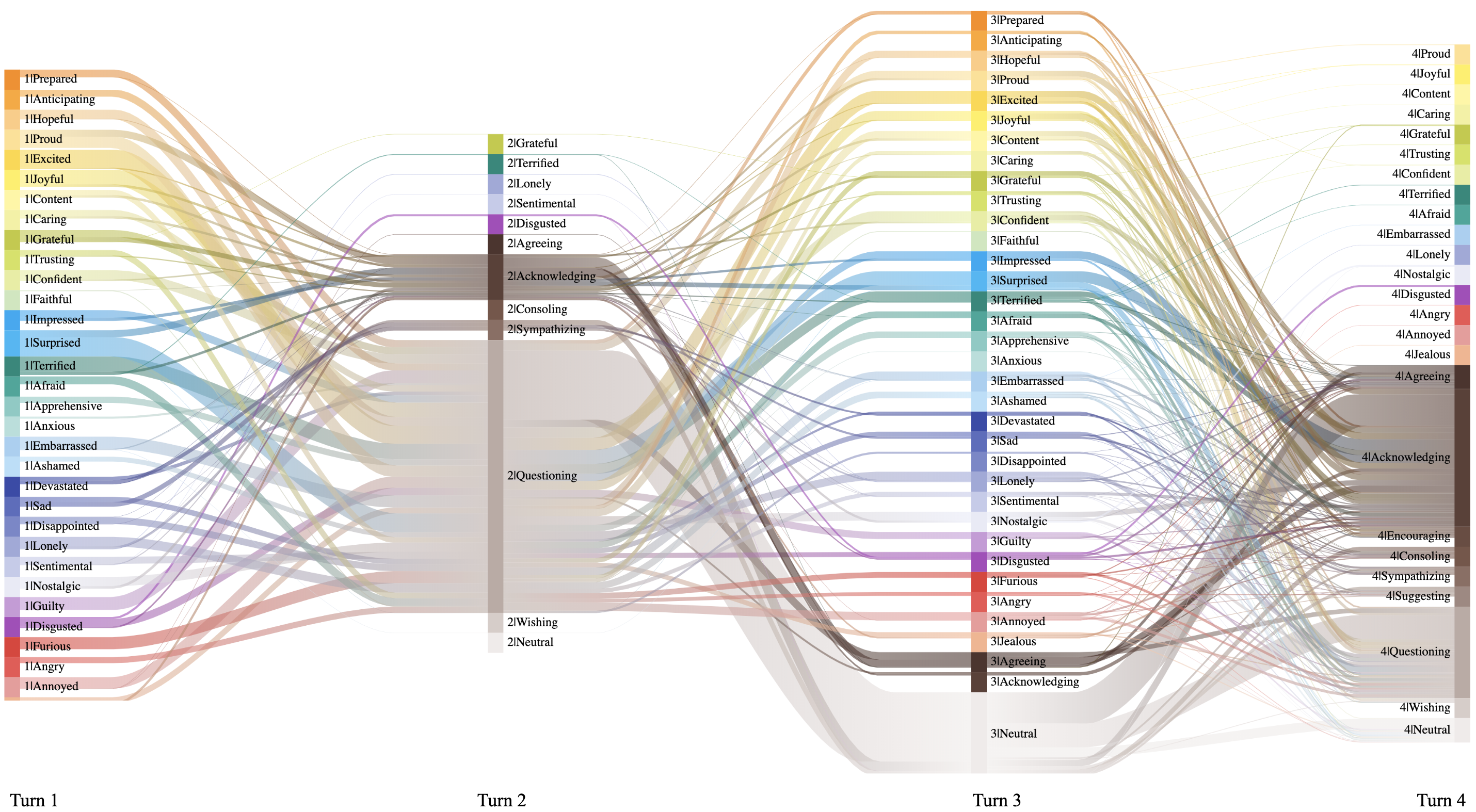}
    \caption{Visualization of the most common emotion-intent flow patterns (having a frequency $\geq 5$) throughout the first four dialogue turns in the EmpatheticDialogues dataset.} 
    \label{fig:temporal}
\end{figure}

Figure \ref{fig:temporal} plots the most frequent emotion-intent flow patterns up to $4$ turns in the dataset. Turns $1$ and $3$ correspond to speaker turns and turns $2$ and $4$ correspond to listener turns. According to the visualization, most emotions experienced by speakers are immediately followed by `questions' as well as expressions of `acknowledgment'. Expressions of `sympathy' immediately follow more negatively intense emotions such as `devastated' and `sad'. Towards the end of dialogues, we can see more expressions of `acknowledgment', `agreement' and `suggesting'. Expressions of `encouragement' and `wishing' can be seen in the case of positive emotional situations and `sympathizing' and `consoling' in the case of negative emotional situations. Another important observation is that towards the end of dialogues, listener utterances become more emotional compared to the beginning, as the speakers elaborate on their emotions. Such situations also reflect scenarios of personal distress---the phenomenon where one is unable to distinguish the emotion of their own from the emotion of another. Dialogue in Table \ref{table:distress} illustrates this phenomenon.  

\begin{table}[t!]
\begin{tabularx}{\textwidth}{|X|}
\hline

S: \textit{Bleh, I just had the worst food ever.} (Disgusted)

L: \textit{What did you eat?} (Questioning)

S: \textit{I was at Mcdonalds and was given a rotten cheese burger. I almost puked after I ate it.} (Disgusted)

L: \textit{Oh gross, makes me never want McDonalds again.} (Disgusted)\\

\hline

\caption{Example conversation that illustrates personal distress towards the end of the dialogue.}\label{table:distress}
\end{tabularx}
\end{table}

Still, they are not as frequent as how listeners choose to empathize healthily instead of making it a distress. But this sheds light on the fact that when the speaker goes on elaborating on his situation, sometimes the listener's ability to distinguish between the speaker's emotion and the emotion of his own may decrease, leading to consequences of personal distress. In the case of intense negative emotions, it can lead to avoidance or a deliberate change of conversation topic. This is a scenario commonly experienced by people engaged in therapeutic and health professions and is described in researches by Singer and Klimecki \shortcite{empathy_and_compassion} and Buechel et al. \shortcite{Modeling_Empathy_and_Distress_in_Reaction_to_News_Stories}. However, in order to verify this observation more solidly, we need a corpus with a larger number of turns per dialogue.


The taxonomy we have developed can be incorporated into the design of social chatbots to gain more controllability and interpretability of the responses generated. It can be achieved by feeding in the conversation history, in which each utterance is tagged with an emotion or an intent label into a neural network that jointly models dialogue intent selection and response generation. Dialogue intent selection module will select the most appropriate intent based on the conversation history we feed in, and the response generation module will generate an appropriate response conditioned on the selected intent label. To help ensure more robustness, it is also possible to repeatedly sample plausible intent labels during training and feed them into the response generation module. The overall goal of modeling chatbots in this manner is to lead the conversation in a healthy and desirable direction with the controllability and interpretability provided by the taxonomy. Moreover, the taxonomy can be used as an annotation scheme to label utterances in other datasets and analyse them in terms of their empathetic quality in the same way described here. It also has implications in distinguishing between multiple forms of empathy---compassion and personal distress, as recognized in psychology and neuroscience fields.

One limitation of this study is the analysis results are highly dependent on the EmpatheticDialogues dataset. For future studies, we intend to curate a much larger empathetic dialogue dataset using a subset of the OpenSubtitles (8 million dialogues) \cite{os_2018}, which will help us develop a more accurate emotion classifier and establish a more general taxonomy. Another limitation is the emotion classifier trained to automatically label utterances in the EmpatheticDialogues dataset is a sentence level classifier, which is unable to accurately distinguish similar utterances whose empathetic label can differ according to the context. We intend to improve our sentence level classifier into a classifier based on dialogue history that will be able to more accurately distinguish such cases. Automated labeling of intents using lexical methods also have the possibility to injure the robustness of the model due to considering only the most indicative n-grams in individual sentences without accounting for the surrounding context. 


\section{Conclusion}
\label{sec:conclusion}

In this paper, we introduced a taxonomy of empathetic response intents capable of supporting automatic empathetic communication in social chitchat. The strategies relying on this taxonomy are essential for a chatbot to engage in prosocial conversations, expressing empathetic concern for its users, and keeping the users engaged. Another significant contribution from our work is to provide analysis on the EmpatheticDialogues corpus after automatically annotating it based on the most frequent intents from our taxonomy and 32 types of emotion categories defined in EmpatheticDialogues. We illustrate the most frequent emotion-intent exchange patterns in the dataset and how they vary temporally over the course of interaction. These results further validate the taxonomy of empathetic listener intents we derived and shed light on the frequent empathetic conversation patterns seen among humans when engaged in social chitchat. We explained how our taxonomy can be utilized in the development of an empathetic chatbot to achieve more controllability and interpretability in the responses generation process. The method described here can also be used as an annotation scheme to label utterances from other datasets and analyse them in terms of their empathetic quality. As future work, we plan on using these findings to develop a social chatbot capable of effectively engaging in empathetic conversations.



\noindent\textbf{Appendix A. Words and phrases most indicative of the empathetic response intents that were used to extract more example listener utterances from the EmpatheticDialogues dataset for training the BERT transformer-based classifier}

\begin{longtable}[ht!]{| p{.17\textwidth} |  p{.77\textwidth} |}


\hline Response intent
& Words and phrases most indicative of the intent
\\ \hline

Agreeing

&

\textit{100\%, exactly, absolutely, definitely, agree, i know, me either, me neither, i understand, i completely understand, me too, that's right, you're right, correct}\\

Acknowledging

&

\textit{it sucks, that sucks, i'd ... too, i would ... too, i feel you, that's splendid, i bet ... was, that's great", that's a good idea, i bet ... can't, that's pretty, i see, it's pretty, can understand, sounds, that would, i would have, must've, cool, nice, awesome}\\

Encouraging

&

\textit{hopefully ... will, i hope ... will, works out for you, i bet ... will, i bet ... 'll, i bet ... can}\\

Consoling

&

\textit{there you go, hopefully ... will, i hope ... will, cheer up, get better, will pass quickly}\\

Sympathizing

&

\textit{i'm sorry, sorry to hear, oh no, bless you, deepest sympathy}\\

Suggesting

&

\textit{maybe, i think ... should, perhaps, why don’t you, you could always, what if}\\

Questioning

&

\textit{what ... ?, why ... ?, when ... ?, where ... ?, how ... ?, are ... ?, is ... ?, did ... ?, do ... ?, does ... ?, have ... ?, has ... ?, had ... ?}\\
Wishing

&

\textit{congratulations, happy birthday, happy anniversary, i wish you, wish you ... !, all the best, good luck}\\\hline

\caption{Words and phrases that are most indicative of the empathetic response intents.}\label{table:response_intents}
\end{longtable}


\noindent\textbf{Appendix B. Example speaker-listener utterance pairs corresponding to the taxonomy of emotion/intent exchanges}

\begin{longtable}[ht!]{| p{.09\textwidth} | p{.15\textwidth} | p{.7\textwidth} |}


\hline Speaker's emotion
& Listener's response emotion/intent & Example utterance-response pairs
\\ \hline


Anticipa

-ting

&

Questioning

\vspace{11.5mm}

Acknowledging&

S: \textit{When tax season came I was in a hurry to get mine done. I was looking forward to a big refund.} (Anticipating)

L: \textit{really? why is that?}	(Questioning)

\vspace{2mm}

S: \textit{I cannot wait for the newest Pokemon game, it looks amazing to me!}	(Anticipating)

L: \textit{Those games do seem fun} (Acknowledging)\vspace{1mm}\\\hline

Joyful

&

Questioning

\vspace{11.5mm}

Acknowledging&

S: \textit{i was happy to see that i was able to get a new pet the other day}	(Joyful)

L: \textit{What pet did you get?} (Questioning)

\vspace{2mm}

S: \textit{I jumped for joy when my baby was born.}	(Joyful)

L: \textit{wow that must have been a huge moment for you}	(Acknowledging)\vspace{1mm}\\\hline

Trusting

& 

Questioning

\vspace{11.5mm}

Acknowledging

&

S: \textit{Man, I let one of my friends take my Benz one day to run some errands. I really thought she would be careful with it.} (Trusting)

L: \textit{Oh, no! Did she damage your car?} (Questioning)

\vspace{2mm}

S: \textit{My therapist was so kind to me, I had to tell her a lot.}	(Trusting)

L: \textit{That's good you have someone that you can talk to about your problems and feelings. I'm sure it helps!} (Acknowledging)\vspace{1mm}\\\hline

Surprised

& 

Questioning

\vspace{7mm}

Acknowledging

\vspace{7mm}

Neutral

& 

S: \textit{I was shocked when i got invited on a random trip}	(Surprised)

L: \textit{Was a happy shocked feeling or a bad one?} (Questioning)

\vspace{2mm}

S: \textit{The other day I found out that my sister is having twins!}	(Surprised)

L: \textit{Oh that's wonderful twins seem really cool.} (Acknowledging)

\vspace{2mm}

S: \textit{No one even knew she was dating anyone until the announcement, so I was very surprised.} (Surprised)

L: \textit{I guess she wanted to keep it a secret for some reason.} (Neutral)\vspace{1mm}\\\hline

Afraid

&

Questioning

\vspace{7mm}

Acknowledging

&

S: \textit{It's so dark and creepy down there.}	(Afraid)

L: \textit{lol.  Do you think there are monsters down there?}	(Questioning)

\vspace{2mm}

S: \textit{It was only off for a little over 2 hours, but I could not find a flashlight and it was so scary.} (Afraid)

L: \textit{That sounds awful!} (Acknowledging)\vspace{1mm}\\\hline

Sad

&

Questioning

\vspace{16mm}

Sympathizing

\vspace{7mm}

Acknowledging

\vspace{6mm}

Agreeing

& 

S: \textit{I feel bad I don't always get to go through bad things and full get healed.}	(Sad)

L: \textit{Do you mean you feel bad that you don't get to go through bad things or that you don't get to be healed?}	(Questioning)

\vspace{2mm}

S: \textit{I was extremely emotional when my dog passed away}	(Sad)

L: \textit{Aww man sorry for your loss, those are the worst.} (Sympathizing)

\vspace{2mm}

S: \textit{My favorite donut shop went out of business.} (Sad)

L: \textit{Ah that's a pity. It really sucks to lose favorite shops.} (Acknowledging)

\vspace{2mm}

S: \textit{I'm sad. My youngest son starts kindergarten tomorrow!} (Sad)

L: \textit{I am sure it is a bittersweet moment.  I can relate myself.}	(Agreeing)\vspace{1mm}\\\hline

Disgusted

&

Questioning

\vspace{11.5mm}

Acknowledging

\vspace{11.5mm}

Disgusted

\vspace{11.5mm}

Agreeing

& 

S: \textit{I am disgusted that so many people voted in favour of Brexit in the UK.}	 (Disgusted)

L: \textit{Why is that?}	(Questioning)

\vspace{2mm}

S: \textit{It was a brand new box of Rice crispies. When I opened it and poured it in my bowl, there were several live bugs.} (Disgusted)

L: \textit{Well that sounds disgusting}	(Acknowledging)

\vspace{2mm}

S: \textit{I was at Mcdonalds and was given a rotten cheese burger. I almost puked after I ate it.} (Disgusted)

L: \textit{Oh gross, makes me never want McDonalds again.} (Disgusted)

\vspace{2mm}

S: \textit{Everytime I see my cat vomit on floor it makes me sick.}	(Disgusted)

L: \textit{i think you have the same attitude like me.}	(Agreeing)\vspace{2mm}\\\hline

Angry

&

Questioning

\vspace{7mm}

Acknowledging

& 

S: \textit{i was upset when i saw someone put a dent in my door} (Angry)

L: \textit{Was this a parking lot?} (Questioning)

\vspace{2mm}

S: \textit{My grandma didn't make my oatmeal right yesterday. I was so mad.}	(Angry)

L: \textit{Oh wow!  You were pretty angry}	(Acknowledging)\vspace{1mm}\\\hline


\caption{Example speaker and listener utterances corresponding to the most common emotion exchanges between speakers and listeners, when the speaker's emotion is one of the Plutchik's 8 basic emotions.}\label{table:emotions}
\end{longtable}

\end{document}